\documentclass[letterpaper, 10pt, conference]{ieeeconf}

\IEEEoverridecommandlockouts
\usepackage{float}

\usepackage{graphicx}
\usepackage{amsmath}
\usepackage{amssymb}
\usepackage{booktabs}
\usepackage{multirow}
\usepackage[table]{xcolor}

\title{\LARGE \bf
OccPlanner: Goal-Aware Occupancy-Conditioned Diffusion Planner for PixelGoal Navigation
}

\author{
Binling Huang$^{*}$, Nianjin Ye, Xi Yang, Liang Hu, Zhou Huang,\\
Shuang Wei, Longrui Yang, Yanchi Chen, and Lanpeng Jia\\[0.5em]
Changhong Intelligent Robot\\
University of Electronic Science and Technology of China
}

\begin{document}

\maketitle
\begingroup

\renewcommand{\thefootnote}{}
\footnotetext{\textsuperscript{*}First author.}
\endgroup

\thispagestyle{empty}
\pagestyle{empty}


\begin{abstract}
PixelGoal navigation specifies targets directly in the agent's camera view, providing a natural interface between high-level visual reasoning and low-level navigation. Depth can lift a visible target pixel into a metric PointGoal, but this estimate becomes unreliable under occlusion or sensor noise. Moreover, a PointGoal alone does not encode traversability or feasible paths around obstacles. We present \textbf{OccPlanner}, a goal-aware occupancy-conditioned diffusion planner that learns complementary egocentric goal and planning-oriented 3D representations through metric target and occupancy prediction, respectively. These representations condition a diffusion trajectory module to generate target-directed, obstacle-aware trajectories. For scalable geometric supervision, we introduce L3ROcc, which converts monocular RGB navigation videos into aligned 3D occupancy and trajectory annotations. We train OccPlanner on L3ROcc-processed InternData-N1 and evaluate it in closed-loop simulation across four unseen InternScenes categories and two goal-distance ranges. Across all eight settings, OccPlanner substantially outperforms existing open-source PixelGoal approaches and achieves competitive performance against PointGoal planners with direct metric-goal inputs.
\end{abstract}

\section{Introduction}
\label{sec:intro}

Embodied visual navigation requires an agent to translate egocentric observations and a goal specification into obstacle-aware motion in unseen environments~\cite{savva2019habitat}. Trajectory generation is particularly challenging because successful navigation requires both goal localization and reasoning about surrounding 3D geometry and traversability under closed-loop execution.

Navigation goals may be specified as metric coordinates, semantic targets, or natural-language instructions~\cite{wijmans2020ddppo,ye2021objectnav,gong2026poinav,anderson2018vln}. With the growing capabilities of vision-language models (VLMs), high-level systems can identify the next location to visit directly in image space~\cite{cai2024pixnav,bao2026goal2pixel}. Pixel-goal navigation provides a direct interface between high-level visual reasoning and low-level motion planning without requiring a pre-built map or an explicit metric goal. However, a pixel does not specify metric target geometry or surrounding traversability. Depth provides metric measurements of visible surfaces, but the planner must still infer target location and obstacle structure from the observation history.

\begin{figure}[t!]
\raggedleft
\includegraphics[width=0.95\linewidth]{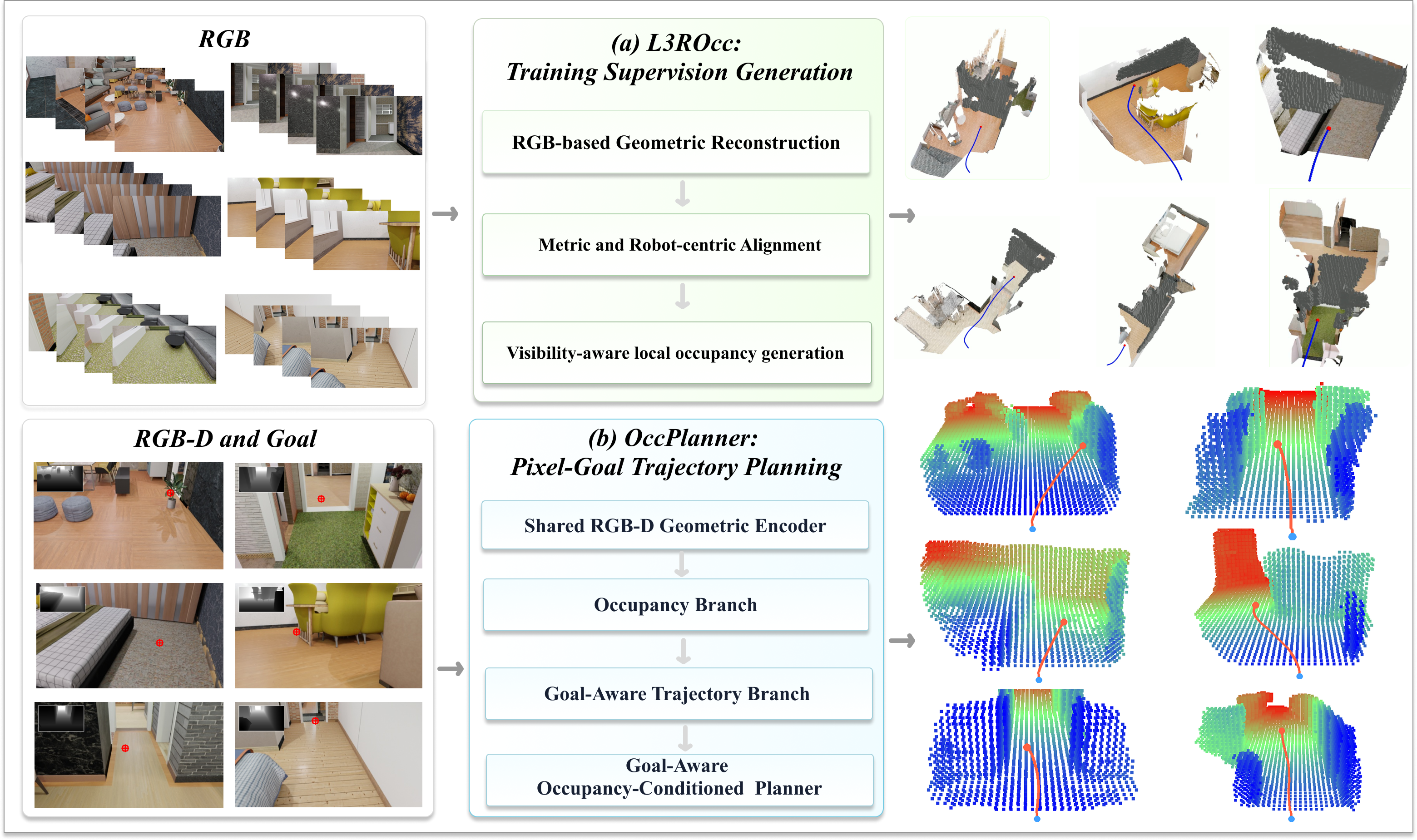}
\caption{\textbf{Overview of OccPlanner and L3ROcc.} (a) L3ROcc generates visibility-aware local 3D occupancy and aligned trajectory annotations from monocular RGB navigation videos. (b) OccPlanner grounds the PixelGoal in RGB-D context and conditions trajectory generation on target-relevant local occupancy.}
\label{fig:teaser}
\end{figure}

Existing approaches leave three gaps. Pixel-goal policies condition on image-space targets or back-project visible goals into metric waypoints, but lack explicit metric grounding from temporal context~\cite{cai2024pixnav,krishnan2026ssmpixnav,bao2026goal2pixel}. Geometry-aware local planners exploit depth or reconstructed 3D structure but assume metric goal inputs~\cite{yang2023iplanner,roth2024viplanner,peng2025logoplanner}. Diffusion navigation policies model multimodal action or trajectory distributions, yet do not explicitly connect PixelGoal grounding with planning-oriented occupancy reasoning~\cite{sridhar2024nomad,cai2026navdp}. Consequently, image-space goal grounding and local 3D reasoning remain disconnected in trajectory generation.

We address this gap with \textbf{OccPlanner}, a \textbf{Goal-Aware Occupancy-Conditioned Diffusion Planner} that jointly reasons about target geometry and local 3D scene structure. Given a PixelGoal and RGB-D history, OccPlanner learns an egocentric metric goal representation through auxiliary target prediction and learns planning-oriented local geometry through occupancy supervision. The resulting features jointly condition a diffusion planner for obstacle-aware trajectory generation~\cite{ho2020denoising,chi2023diffusionpolicy}.

To provide geometric supervision at scale, we introduce \textbf{L3ROcc} (\textbf{L}ocal \textbf{3D} \textbf{R}econstruction with \textbf{Occ}upancy), which converts monocular RGB navigation videos into temporally consistent local geometric annotations. Built on $\pi^3$~\cite{wang2026pi3}, L3ROcc combines multi-frame geometric reconstruction with voxel-space visibility reasoning to distinguish occupied, observed-free, and unknown regions.

We evaluate OccPlanner on about 5{,}000 valid closed-loop episodes across 60 unseen scenes from InternScenes~\cite{InternScenes}, covering diverse scene layouts and goal distances. OccPlanner outperforms the PixelGoal baselines across all eight scene--distance settings and remains within 3.50 percentage points of the strongest PointGoal reference on the long-range split. Extensive ablations validate the proposed goal and occupancy conditioning mechanisms. Closed-loop experiments on a Unitree Go2 further assess real-world transfer.

Our contributions are threefold:
\begin{itemize}
\item We formulate PixelGoal navigation as a geometry-aware trajectory planning problem and introduce \textbf{OccPlanner}, which grounds image-space goals in metric 3D geometry for obstacle-aware planning.
\item We develop a \textbf{goal-aware occupancy-conditioning mechanism} that provides target-relevant local geometry directly to a diffusion trajectory policy.
\item We introduce \textbf{L3ROcc},  a scalable pipeline that generates aligned trajectory and visibility-aware local occupancy supervision from monocular RGB navigation videos, enabling occupancy-aware planner training without manually annotated 3D occupancy.
\end{itemize}

\section{Related Work}
\label{sec:related_work}




\subsection{Navigation Representations and Goals}

Learning-based navigation has explored different representations for spatial reasoning and long-horizon planning. 
Spatial memory and learned mapping provide structured scene representations for navigation~\cite{henriques2018mapnet,chaplot2020activeneuralslam}, 
while topological representations support long-range navigation through sparse connectivity rather than dense metric maps~\cite{savinov2018semiparametric,chaplot2020neuraltopological,shah2021ving}. 
Learning from large-scale robot experience further improves generalization across environments and embodiments~\cite{shah2023gnm,shah2023vint}, 
while offline reinforcement learning enables prior navigation experience to be reused for different behaviors~\cite{shah2023revind}.

Navigation goals have also become increasingly flexible. 
PointGoal navigation specifies a metric target~\cite{wijmans2020ddppo}, 
whereas ObjectGoal navigation specifies a semantic object category~\cite{ye2021objectnav}. 
Point of Interest Goal navigation extends goal specification to semantically meaningful locations~\cite{gong2026poinav}, 
while vision-and-language navigation expresses navigation intent through natural-language instructions~\cite{anderson2018vln,chen2022duet}. 
PixelGoal navigation instead specifies the target directly in the image plane~\cite{cai2024pixnav,krishnan2026ssmpixnav}. 
Despite its simple and intuitive interface, translating image-space intent into executable motion still requires metric target grounding and local geometric reasoning.


\subsection{PixelGoal Navigation and Local Planning}

PixelGoal navigation has been explored as a direct image-space interface for goal specification~\cite{cai2024pixnav,krishnan2026ssmpixnav}. 
Recent work further connects semantic goals with pixel-level targets and spatial waypoints~\cite{bao2026goal2pixel}. 
Meanwhile, learning-based local planners generate trajectories from geometric observations and spatial goals~\cite{yang2023iplanner,roth2024viplanner}. 
However, these two directions remain largely separated: PixelGoal methods focus on image-space goal specification, while local planners typically assume an explicit spatial goal. 
OccPlanner bridges this gap by learning metric target grounding and local geometry for obstacle-aware trajectory generation.

\section{Problem Formulation}
\label{sec:formulation}

{At navigation step $t$, the agent receives a history
$\mathcal{H}_t=\{(\mathbf{I}_i,\mathbf{D}_i)\}_{i=t-T+1}^{t}$,
where $\mathbf{I}_i\in\mathbb{R}^{H\times W\times3}$ and
$\mathbf{D}_i\in\mathbb{R}^{H\times W}$, together with a normalized PixelGoal input
$\mathbf{g}_{\mathrm{pix},t}=[u_t,v_t]\in[0,1]^2$. The agent predicts
$\mathbf{A}_t=\{\Delta\mathbf{a}_{\ell}\}_{\ell=1}^{L}$, where
$\Delta\mathbf{a}_{\ell}=(\Delta x_{\ell},\Delta y_{\ell},\Delta\theta_{\ell})$
is the difference between consecutive trajectory states expressed in the fixed
egocentric frame at step $t$:
\begin{equation}
\hat{\mathbf{A}}_t
\sim
p_{\theta}
\left(
\cdot
\mid
\mathcal{H}_t,\mathbf{g}_{\mathrm{pix},t}
\right).
\label{eq:problem_policy}
\end{equation}
The local trajectory is recovered as
$\mathbf{p}_{\ell}=\mathbf{p}_{0}+\sum_{j=1}^{\ell}\Delta\mathbf{a}_{j}$,
with $\mathbf{p}_{0}=\mathbf{0}$. Unlike a metric PointGoal, a PixelGoal 
specifies only image-space intent.  We therefore ground it as an egocentric goal 
$\mathbf{g}_{\mathrm{ego}}=(x_g,y_g,\phi_g)$, where $\phi_g$ denotes the terminal heading of the reference trajectory,
providing the planner with an explicit egocentric representation of where to navigate.
}

\section{Methodology}
\label{methodology}

\subsection{Overview}

Our methodology comprises L3ROcc and OccPlanner. L3ROcc derives scalable local geometric supervision from monocular navigation videos. Using this supervision, OccPlanner learns planning-oriented local geometry from RGB-D observations. Meanwhile, it grounds the PixelGoal in an egocentric metric representation. The resulting goal and occupancy representations jointly condition a diffusion trajectory policy.

\subsection{L3ROcc: Scalable Geometric Supervision}
\label{sec:l3rocc}

\begin{figure*}[!t]
    \centering
    \includegraphics[width=0.95\textwidth]{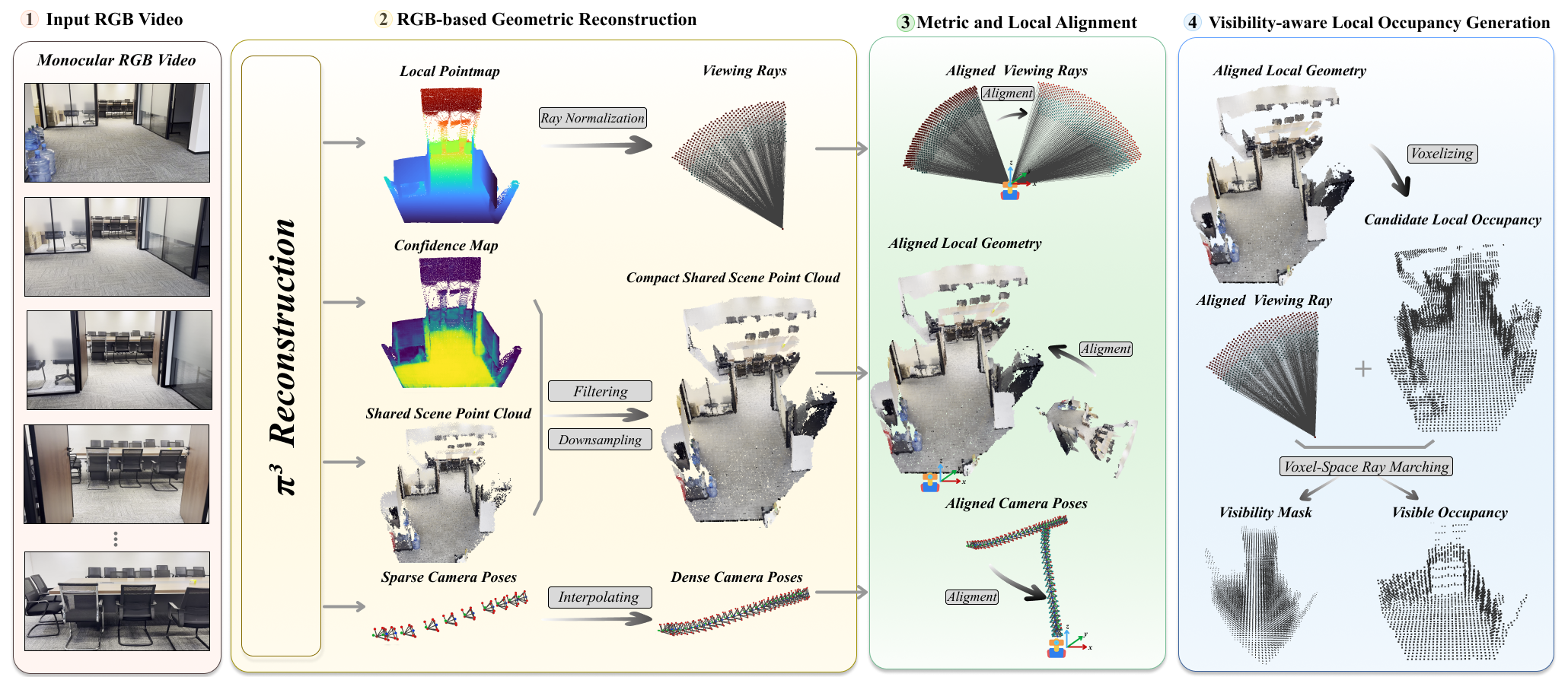}
    \caption{\textbf{L3ROcc data-generation pipeline.} Given a monocular RGB navigation video, L3ROcc reconstructs shared scene geometry and camera motion. After scale and local alignment, ray-based visibility reasoning produces occupancy labels, while the aligned poses form the camera trajectory.}
    \label{fig:l3rocc_pipeline}
\end{figure*}

L3ROcc converts monocular RGB navigation videos into visibility-aware local 3D occupancy and aligned trajectory annotations. As shown in Fig.~\ref{fig:l3rocc_pipeline}, it reconstructs shared scene geometry and camera motion from multiple views. Visibility reasoning produces occupancy labels, while the aligned camera poses provide trajectory supervision.

\subsubsection{RGB-Based Geometric Reconstruction}

Given a monocular RGB video $\mathcal{V}=\{\mathbf{I}_t\}_{t=1}^{N}$, we sample $N_s$ frames at a fixed interval and process them with $\pi^3$~\cite{wang2026pi3}. The model predicts local pointmaps, confidence maps, shared-frame 3D points, and sparse camera poses. We remove low-confidence points and depth-edge artifacts. Voxel downsampling then produces a compact shared scene point cloud $\mathcal{P}^{g}$.

We interpolate the sparse camera poses to all $N$ timestamps. Translation is interpolated using cubic splines~\cite{deboor1978practical}, while rotation uses spherical linear interpolation (SLERP)~\cite{shoemake1985animating}. This yields dense camera poses $\{\mathbf{T}^{g}_{c,t}\}_{t=1}^{N}$, where $\mathbf{T}^{g}_{c,t}$ maps the camera frame to the shared reconstruction frame. We construct a fixed viewing-ray template $\mathcal{R}^{c}$ in the camera frame from the first local pointmap and reuse it at every timestamp for visibility reasoning.

\subsubsection{Metric and Local Alignment}

\begin{figure}[t]
\centering
\includegraphics[width=\linewidth]{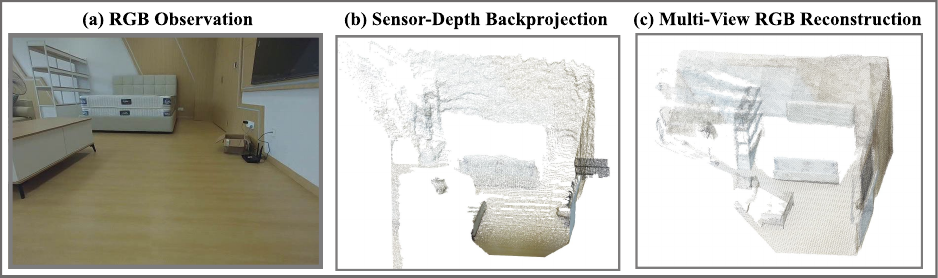}
\caption{\textbf{Real-world geometric reconstruction.} Comparison between sensor-depth backprojection and multi-view RGB reconstruction.}
\label{fig:real_reconstruction}
\end{figure}

Rather than directly voxelizing sensor depth, L3ROcc reconstructs scene geometry from multiple RGB views. Fig.~\ref{fig:real_reconstruction} shows that multi-view reconstruction yields more coherent scene geometry than sparse and noisy sensor-depth backprojection. We resolve the reconstruction's global scale ambiguity using available metric measurements~\cite{eigen2014depth}. A single scale factor per sequence is estimated from a reference camera trajectory or by aligning valid sensor and reconstructed depths.

The selected scale is applied to the shared point cloud and camera translations. At each timestamp, we use the dense camera pose and calibrated camera-to-base matrix to transform the point cloud and viewing rays into the robot frame, yielding the local point cloud $\mathcal{P}^{b}_{t}$ and viewing-ray set $\mathcal{R}^{b}_{t}$. The aligned camera poses are also used to derive the future motion labels $\mathbf{A}_t$.

\subsubsection{Visibility-Aware Local Occupancy Generation}

We voxelize the point cloud $\mathcal{P}^{b}_{t}$ into candidate occupancy $\widetilde{\mathbf{O}}_{t}$ and cast the rays $\mathcal{R}^{b}_{t}$ through the grid following Occ3D~\cite{tian2023occ3d}. Along each ray, voxels before the first hit are observed free, the hit is visible occupied, and voxels behind it are unobserved; if no hit occurs, all traversed voxels are observed free. This yields binary visible occupancy $\mathbf{O}_t$ and a visibility mask $\mathbf{M}_t$, where $\mathbf{M}_t=1$ and $\mathbf{M}_t=0$ denote observed and unobserved voxels, respectively.

\begin{figure*}[!t]
    \centering
    \includegraphics[width=0.9\textwidth]{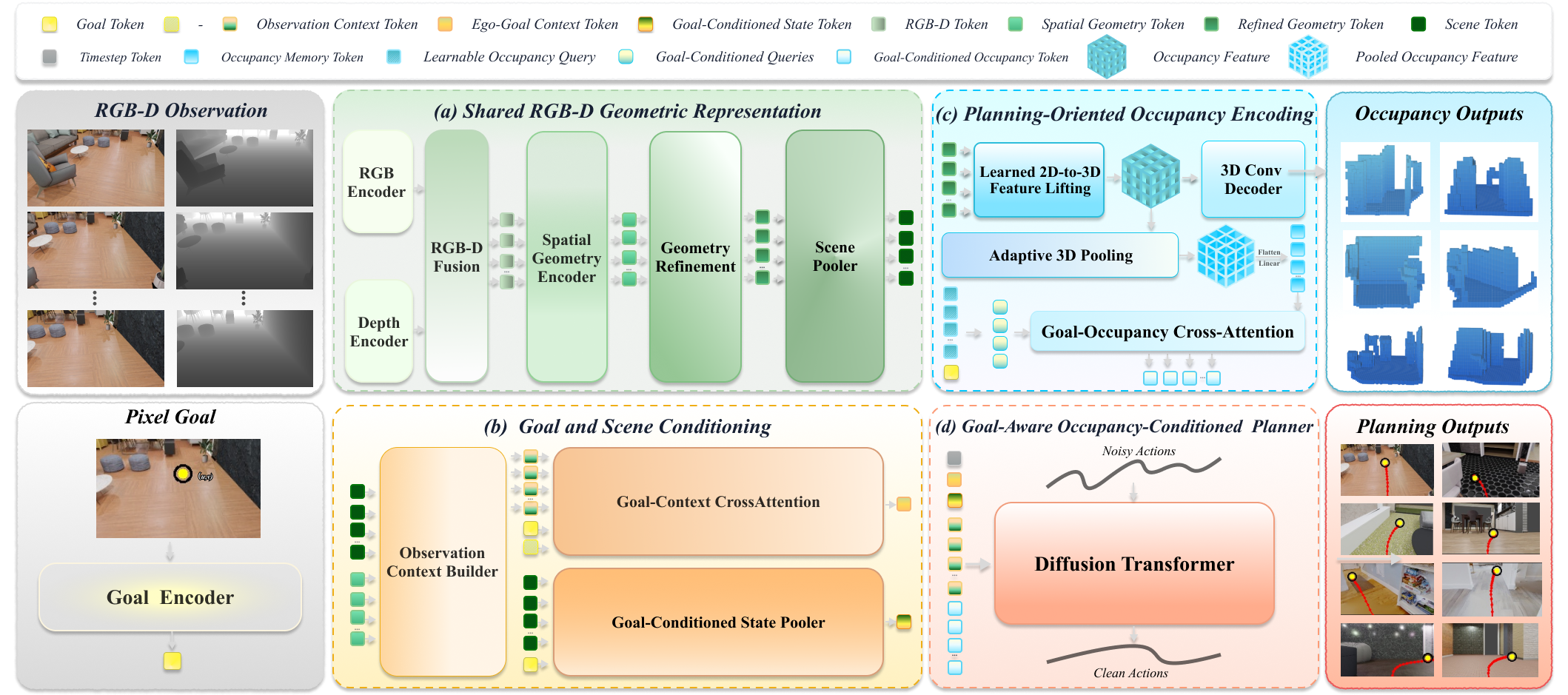}
    \caption{\textbf{OccPlanner.}
    \textbf{(a)} Shared RGB-D encoding extracts spatial geometry and scene features.
    \textbf{(b)} Goal and scene conditioning grounds the PixelGoal in an egocentric target representation.
    \textbf{(c)} Planning-oriented occupancy encoding reconstructs local geometry and extracts target-relevant occupancy features.
    \textbf{(d)} The grounded goal and occupancy features jointly condition diffusion-based trajectory generation.}
    \label{fig:occplanner_pipeline}
\end{figure*}

\subsection{OccPlanner}
\label{sec:occplanner}

As illustrated in Fig.~\ref{fig:occplanner_pipeline}, an RGB-D encoder produces shared geometry tokens from the observation history. The goal branch grounds the PixelGoal against this context. In parallel, the occupancy branch lifts the current geometry into a 3D feature volume and pools it into occupancy memory. Goal-conditioned queries read this memory, and the resulting goal and occupancy features condition the diffusion planner.

\subsubsection{Shared RGB-D Geometric Representation}
\label{sec:shared_geometry}

We index the frames in $\mathcal{H}_t$ by $i=1,\ldots,T$, where $i=T$ denotes the current frame. For each frame, an RGB encoder adopting the DINOv2-S/14 architecture~\cite{oquab2024dinov2} and a compact convolutional depth encoder extract spatially aligned tokens. The RGB-D fusion module concatenates and projects these tokens into a common space. A spatial geometry encoder processes the fused tokens to obtain spatial geometry tokens $\mathbf{Z}^{\mathrm{spa}}_i$, and a geometry refinement module produces refined geometry tokens $\mathbf{Z}^{\mathrm{ref}}_i$. A scene pooler summarizes each $\mathbf{Z}^{\mathrm{ref}}_i$ as a scene token $\mathbf{s}_i$.

\subsubsection{Goal and Scene Conditioning}
\label{sec:goal_scene_conditioning}

A PixelGoal identifies a location in the image but does not specify its metric position in the robot frame. We first construct an observation context that retains both historical scene information and current spatial detail.
The observation context builder projects and compresses the current spatial geometry tokens $\mathbf{Z}^{\mathrm{spa}}_T$ into an observation context token $\mathbf{u}_T$. It then concatenates $\mathbf{u}_T$ with the frame-wise scene tokens: $\mathbf{U}_{\mathrm{ctx}}=[\mathbf{s}_1;\ldots;\mathbf{s}_T;\mathbf{u}_T].$

The goal encoder maps $\mathbf{g}_{\mathrm{pix},t}$ to a goal token $\mathbf{z}_g$ using multi-frequency Fourier features~\cite{tancik2020fourier} followed by an MLP. The goal-context cross-attention module uses $\mathbf{z}_g$ to condition a learnable goal query $\mathbf{q}_{\mathrm{goal}}$, which reads $\mathbf{U}_{\mathrm{ctx}}$ and produces an ego-goal context token $\mathbf{z}_e$. In parallel, the goal-conditioned state pooler summarizes the scene tokens and $\mathbf{z}_g$ into a state token $\mathbf{z}_s$:
\begin{equation}
\begin{aligned}
\mathbf{z}_e
&=
\operatorname{GoalContextAttn}
\left(
\mathbf{q}_{\mathrm{goal}}+\mathbf{z}_g;
\mathbf{U}_{\mathrm{ctx}}
\right),\\
\mathbf{z}_s
&=
\operatorname{StatePool}
\left(
[\mathbf{s}_1;\ldots;\mathbf{s}_T;\mathbf{z}_g]
\right).
\end{aligned}
\label{eq:goal_conditioning}
\end{equation}
During training, an auxiliary head predicts $\hat{\mathbf{g}}_{\mathrm{ego}}$ from $\mathbf{z}_e$, supervised by the expert trajectory expressed in the fixed egocentric frame at step $t$. This auxiliary task grounds the PixelGoal in an egocentric metric position and reference heading, the planner uses $\mathbf{z}_e$ rather than the predicted coordinates.

\subsubsection{Planning-Oriented Occupancy Encoding}
\label{sec:planning_occupancy}

Metric goal grounding provides target guidance but does not capture the surrounding obstacle geometry. Although depth provides visible surface measurements, it does not explicitly require the learned representation to preserve obstacle layout. We therefore use occupancy prediction to organize local geometry on a fixed grid in the robot frame and supervise the shared representation.

Learned 2D-to-3D feature lifting maps the current refined geometry tokens $\mathbf{Z}^{\mathrm{ref}}_T$ to an occupancy feature volume $\mathbf{F}_{\mathrm{occ}}$. A 3D convolutional decoder predicts visible-occupancy logits $\hat{\mathbf{O}}_t$ under the L3ROcc supervision described in Sec.~\ref{sec:l3rocc}, encouraging $\mathbf{F}_{\mathrm{occ}}$ to encode
local obstacle structure.

When used only as an auxiliary prediction objective, occupancy supervision influences trajectory generation indirectly through the shared representation. We therefore expose $\mathbf{F}_{\mathrm{occ}}$ directly to the planner. Adaptive 3D pooling and projection produce occupancy memory tokens $\mathbf{U}_{\mathrm{occ}}$. The projected goal token is added to each of $K$ learnable queries, which attend to this memory:
\begin{equation}
\mathbf{Z}_{\mathrm{occ}}
=
\operatorname{GoalOccAttn}
\left(
\mathbf{Q}_{\mathrm{occ}}+\mathbf{W}_g\mathbf{z}_g;
\mathbf{U}_{\mathrm{occ}}
\right).
\label{eq:occupancy_conditioning}
\end{equation}
The resulting tokens provide target-relevant local geometry to the planner. This readout remains differentiable, allowing the trajectory objective to shape the occupancy representation.

\subsubsection{Goal-Aware Occupancy-Conditioned Planner}
\label{sec:planner}

The goal and occupancy pathways condition the planner in parallel. Following DDPM~\cite{ho2020denoising} and Diffusion Policy~\cite{chi2023diffusionpolicy}, we formulate trajectory generation as a conditional denoising process. At diffusion step $k$, the noisy action tokens $\mathbf{A}^{(k)}_t$ serve as queries, while the conditioning sequence is
\begin{equation}
\begin{aligned}
\mathbf{C}_k
&=
[\mathbf{e}_k;
 \mathbf{z}_e;
 \mathbf{z}_s;
 \mathbf{U}_{\mathrm{ctx}};
 \mathbf{Z}_{\mathrm{occ}}],\\
\hat{\boldsymbol{\epsilon}}
&=
\epsilon_\theta
\left(
\mathbf{A}^{(k)}_t;
\mathbf{C}_k
\right).
\end{aligned}
\label{eq:joint_planning}
\end{equation}
where $\mathbf{e}_k$ is the diffusion timestep embedding and $\epsilon_\theta$ denotes the diffusion transformer and its noise prediction head. The ego-goal context token $\mathbf{z}_e$ provides metric target information, while the goal-conditioned occupancy tokens $\mathbf{Z}_{\mathrm{occ}}$ provide target-relevant local geometry. Reverse denoising produces the motion increments $\mathbf{A}_t$ defined in Sec.~\ref{sec:formulation}.

We jointly optimize trajectory denoising, trajectory reprojection, ego-goal regression, and occupancy prediction:
\begin{equation}
\mathcal{L} =
\lambda_{\mathrm{traj}}\mathcal{L}_{\mathrm{traj}}
+\lambda_{\mathrm{uv}}\mathcal{L}_{\mathrm{uv}}
+\lambda_{\mathrm{ego}}\mathcal{L}_{\mathrm{ego}}
+\lambda_{\mathrm{occ}}\mathcal{L}_{\mathrm{occ}}.
\label{eq:total_loss}
\end{equation}
$\mathcal{L}_{\mathrm{traj}}$ applies Smooth L1 loss between the predicted and sampled noise, while $\mathcal{L}_{\mathrm{ego}}$ applies Smooth L1 loss between the predicted and ground-truth egocentric goals. The $\mathcal{L}_{\mathrm{uv}}$ applies Smooth L1 loss between the predicted and ground-truth trajectories projected onto the current image frame. $\mathcal{L}_{\mathrm{occ}}$ applies focal loss~\cite{lin2017focal} to $\mathbf{O}_t$: visible occupied voxels are positive, while observed-free and unobserved voxels are negative. In our setting, we find that grouping observed-free and unobserved voxels provides effective geometric supervision for trajectory planning.

\begin{table*}[!t]
\centering
\caption{Closed-loop navigation results in unseen InternScenes environments. All methods use the same episodes within each split. SR/SPL are reported in percent and DTG in meters; the best PointGoal and PixelGoal results are bolded. NavDP-PixelGoal denotes our PixelGoal adaptation of the published PointGoal NavDP.}
\label{tab:sim-main}
\scriptsize
\setlength{\tabcolsep}{1.2pt}
\renewcommand{\arraystretch}{1.10}
\begin{tabular*}{0.92\textwidth}{@{\extracolsep{\fill}}ll*{12}{c}@{}}
\toprule
\multirow{2}{*}{\textbf{Goal Input}} & \multirow{2}{*}{\textbf{Method}}
& \multicolumn{3}{c}{\textbf{Home}}
& \multicolumn{3}{c}{\textbf{Commercial}}
& \multicolumn{3}{c}{\textbf{Cluttered Easy}}
& \multicolumn{3}{c}{\textbf{Cluttered Hard}} \\
\cmidrule(lr){3-5}\cmidrule(lr){6-8}\cmidrule(lr){9-11}\cmidrule(lr){12-14}
& & SR$\uparrow$ & SPL$\uparrow$ & DTG$\downarrow$
& SR$\uparrow$ & SPL$\uparrow$ & DTG$\downarrow$
& SR$\uparrow$ & SPL$\uparrow$ & DTG$\downarrow$
& SR$\uparrow$ & SPL$\uparrow$ & DTG$\downarrow$ \\
\midrule
\rowcolor{gray!10}
\multicolumn{14}{c}{\textbf{Medium Split (3--5\,m)}} \\
\multirow[t]{3}{*}{PointGoal}
& NavDP
& 49.81 & 49.59 & 0.917 & 52.09 & 52.07 & 0.878
& 94.33 & 94.29 & 0.358 & 90.85 & 90.74 & \textbf{0.343} \\
& iPlanner
& \textbf{65.92} & \textbf{65.72} & \textbf{0.757}
& \textbf{72.41} & \textbf{72.03} & \textbf{0.612}
& \textbf{97.07} & \textbf{95.89} & \textbf{0.181}
& \textbf{91.83} & \textbf{91.16} & 0.360 \\
& ViPlanner
& 49.14 & 48.28 & 0.940 & 54.03 & 52.82 & 0.813
& 63.43 & 63.38 & 0.598 & 52.97 & 52.87 & 0.802 \\
\addlinespace[1pt]
StartGoal
& LoGoPlanner
& 38.58 & 38.28 & 1.288 & 40.55 & 40.39 & 1.382
& 45.52 & 45.32 & 3.342 & 38.94 & 38.61 & 3.150 \\
\addlinespace[1pt]
\multirow[t]{3}{*}{PixelGoal}
& NavDP-PixelGoal
& 10.08 & 9.73 & 1.517 & 6.58 & 6.37 & 1.577
& 30.49 & 30.36 & 1.513 & 25.96 & 25.93 & 1.617 \\
& PixNav
& 1.23 & 1.15 & 2.733 & 2.72 & 2.59 & 2.492
& 10.19 & 10.19 & 2.474 & 10.26 & 10.26 & 2.524 \\
\rowcolor{gray!5}
& \textbf{OccPlanner (Ours)}
& \textbf{62.26} & \textbf{61.36} & \textbf{0.856}
& \textbf{66.74} & \textbf{65.84} & \textbf{0.741}
& \textbf{95.16} & \textbf{93.21} & \textbf{0.199}
& \textbf{94.10} & \textbf{92.53} & \textbf{0.213} \\
\addlinespace[2pt]
\rowcolor{gray!10}
\multicolumn{14}{c}{\textbf{Hard Split (5--8\,m)}} \\
\multirow[t]{3}{*}{PointGoal}
& NavDP
& 41.15 & 40.76 & \textbf{1.410}
& 39.31 & 39.13 & \textbf{1.300}
& 92.03 & 91.97 & 0.435 & 85.03 & 84.83 & 0.533 \\
& iPlanner
& \textbf{49.70} & \textbf{49.50} & 1.523
& \textbf{52.16} & \textbf{52.00} & 1.461
& \textbf{96.11} & \textbf{95.78} & \textbf{0.278}
& \textbf{95.88} & \textbf{95.01} & \textbf{0.265} \\
& ViPlanner
& 36.98 & 36.51 & 1.577 & 37.17 & 36.49 & 1.455
& 63.01 & 62.90 & 0.641 & 52.55 & 52.38 & 0.786 \\
\addlinespace[1pt]
StartGoal
& LoGoPlanner
& 20.97 & 20.78 & 1.916 & 20.95 & 20.79 & 1.965
& 37.37 & 36.98 & 3.736 & 40.40 & 39.84 & 4.286 \\
\addlinespace[1pt]
\multirow[t]{3}{*}{PixelGoal}
& NavDP-PixelGoal
& 9.46 & 9.24 & 2.202 & 8.62 & 8.35 & 2.179
& 18.45 & 18.39 & 2.247 & 20.12 & 20.00 & 2.423 \\
& PixNav
& 0.09 & 0.09 & 4.805 & 0.18 & 0.17 & 4.463
& 2.01 & 2.01 & 4.374 & 3.17 & 3.17 & 4.467 \\
\rowcolor{gray!5}
& \textbf{OccPlanner (Ours)}
& \textbf{47.83} & \textbf{47.11} & \textbf{1.673}
& \textbf{45.81} & \textbf{45.22} & \textbf{1.615}
& \textbf{94.78} & \textbf{92.81} & \textbf{0.328}
& \textbf{91.44} & \textbf{89.44} & \textbf{0.541} \\
\bottomrule
\end{tabular*}
\end{table*}

\section{Experiments}
\label{sec:experiments}

\subsection{Experimental Setup}

\begin{figure}[!t]
\centering
\includegraphics[width=\linewidth]{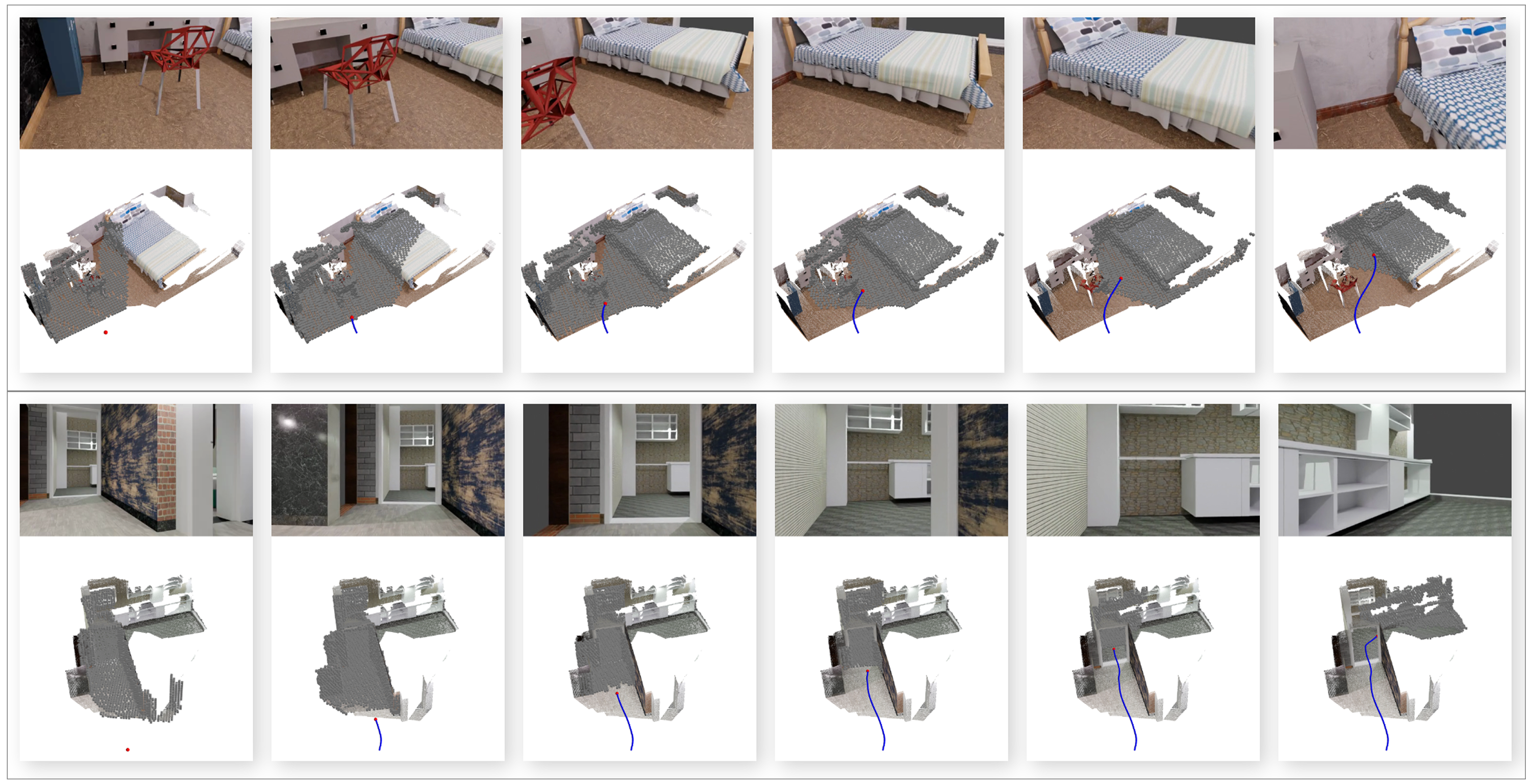}
\caption{\textbf{L3ROcc supervision.} Visibility-aware local occupancy annotations aligned with expert trajectories.}
\label{fig:l3rocc_results}
\end{figure}

\noindent\textbf{Training Data and Annotations.}
We train OccPlanner on over 200{,}000 expert trajectories from InternData-N1~\cite{interndata_n1}, collected in diverse simulated indoor environments using a differential-drive robot with a top-mounted RGB-D camera. Robot height and camera pitch are randomized during data collection. L3ROcc provides occupancy annotations aligned with these trajectories. At each training step, the navigation target is projected into the current image to obtain the PixelGoal label. Its egocentric metric position and reference terminal heading provide the ego-goal supervision. Representative annotations are shown in Fig.~\ref{fig:l3rocc_results}.

\noindent\textbf{Implementation Details.} We train OccPlanner for 30 epochs on four NVIDIA H100 GPUs using Adam, a global batch size of 32, bfloat16 precision, and gradient clipping at 1.0. The learning rate decreases linearly from $1\times10^{-4}$ to $5\times10^{-5}$ over the first 10{,}000 steps and remains fixed thereafter. With 120.0M trainable parameters, OccPlanner predicts 24 motion increments from eight $224\times224$ RGB-D frames through 10 denoising steps and takes 0.104\,s per inference on a single NVIDIA RTX 4090. L3ROcc produces a $100\times140\times60$ occupancy grid at 4\,cm resolution, covering 4.0\,m laterally, 5.6\,m forward, and 2.4\,m vertically. We max-pool it to $50\times70\times30$ (8\,cm) for training, use $K=8$ occupancy queries, and set $(\lambda_{\mathrm{traj}},\lambda_{\mathrm{uv}},\lambda_{\mathrm{ego}},\lambda_{\mathrm{occ}})=(0.8,1.0,0.25,1.0)$.

\noindent\textbf{Evaluation Setup and Metrics.}
We evaluate closed-loop navigation with a Clearpath Dingo robot in NVIDIA Isaac Sim on 60 unseen InternScenes~\cite{InternScenes}: 20 Home, 20 Commercial, 10 Cluttered Easy, and 10 Cluttered Hard scenes. We sample 50 start--goal pairs per scene in each geodesic distance range, 3--5\,m and 5--8\,m, yielding about 5{,}000 valid candidate episodes. We exclude episodes affected by invalid simulator outputs or shared physics artifacts. This shared filter leaves 2{,}437 medium-range and 2{,}672 long-range episodes. Targets lie on navigable ground and are initially visible but may later become occluded. Under the reprojected PixelGoal protocol, the simulator projects the fixed target into the current image at each step. When the target is out of view, the planner produces no new trajectory, and MPC continues executing the last valid one. We report success rate (SR), success weighted by path length (SPL), and final distance to goal (DTG), with success defined as stopping within 0.5\,m of the target. Open-loop evaluation reports average and final displacement errors (ADE and FDE), average and final heading errors (AHE and FHE), and visible-occupancy intersection over union (IoU).

\begin{figure*}[!t]
\centering
\includegraphics[width=0.92\textwidth]{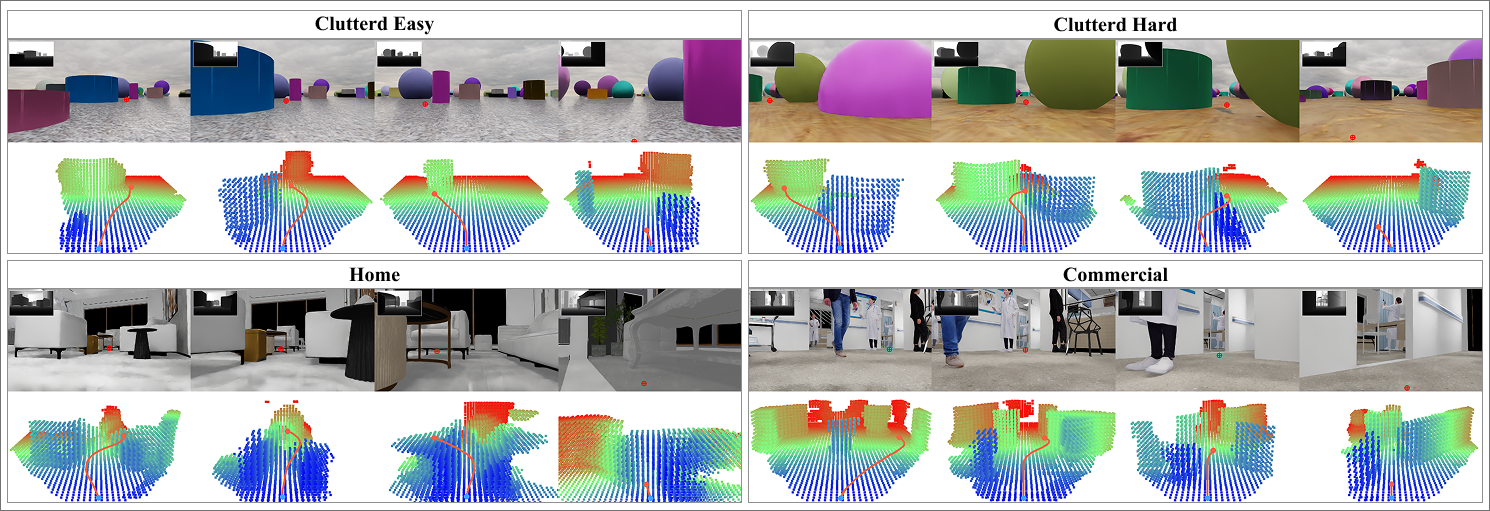}
\caption{\textbf{Qualitative planning results.} Predicted local occupancy and trajectories across unseen scene categories.}
\label{fig:planner_qualitative_results}
\end{figure*}

\begin{figure*}[!t]
\centering
\includegraphics[width=0.92\textwidth]{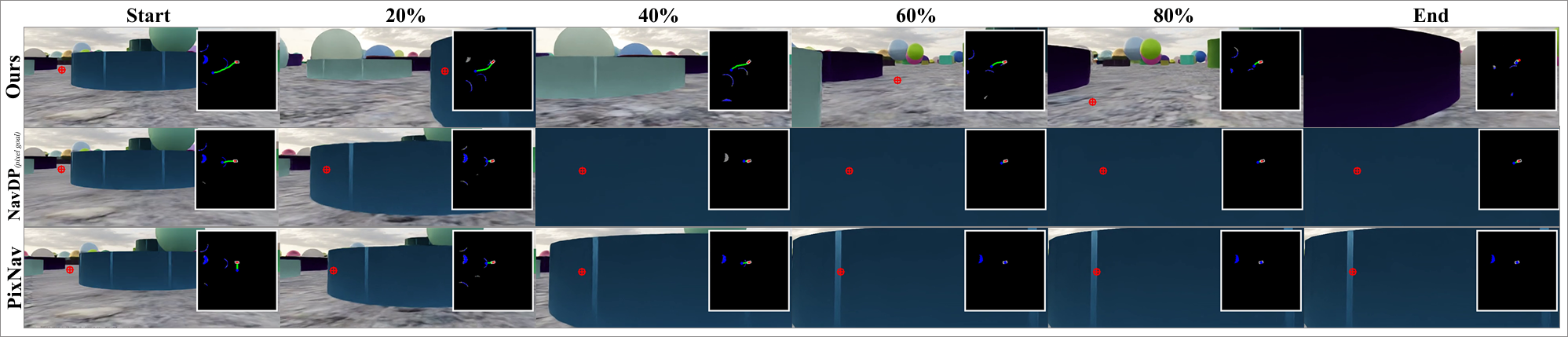}
\caption{\textbf{Closed-loop PixelGoal comparison.} OccPlanner passes the intervening obstacle, while NavDP-PixelGoal and PixNav remain blocked.}
\label{fig:pixelgoal_comparison}
\end{figure*}

\subsection{Experimental Analysis}
\label{sec:experimental_analysis}

We evaluate OccPlanner through closed-loop navigation performance, component ablations, analyses across scene complexities and goal distances, and real-world deployment.

\noindent\textbf{Comparison with Existing Planners.}
We compare OccPlanner with two image-space baselines and four metric-goal references. NavDP-PixelGoal uses the PixelGoal interface and auxiliary metric regression provided in the official NavDP implementation~\cite{cai2026navdp}, although the final publication does not report a standalone PixelGoal benchmark. OccPlanner and NavDP-PixelGoal follow the reprojected PixelGoal protocol described above, while PixNav~\cite{cai2024pixnav} tracks the initially specified pixel under its native protocol. PointGoal NavDP, iPlanner~\cite{yang2023iplanner}, and ViPlanner~\cite{roth2024viplanner} receive the current relative metric goal, whereas LoGoPlanner~\cite{peng2025logoplanner} receives a fixed StartGoal in the initial robot frame.

Replacing NavDP's native PointGoal input with a PixelGoal reduces mean SR by 53.49 and 50.22 points on the medium and hard splits, despite auxiliary metric regression. This comparison confirms that a separately encoded pixel does not provide sufficient metric guidance for execution. OccPlanner addresses this gap by grounding the target against the observation context and conditioning the planner on structured local occupancy. It outperforms both image-space baselines across all settings, with gains of 76.33 and 71.32 SR points over NavDP-PixelGoal in long-range Cluttered Easy and Cluttered Hard. PixNav uses RGB history to track the initially specified pixel and predicts discrete actions. It performs better on short-range goals but degrades as the goal distance increases. This limitation is more pronounced in our 5--8\,m scenes with intervening obstacles.

OccPlanner further outperforms PointGoal NavDP, ViPlanner, and LoGoPlanner across all settings, and remains within 2.24 and 3.50 mean SR points of iPlanner on the two splits. iPlanner retains a small advantage because it receives the current PointGoal directly and learns paths through geometric, goal, and collision costs. LoGoPlanner instead estimates the current target displacement from a fixed StartGoal, so its performance also depends on implicit localization. Overall, OccPlanner recovers sufficiently accurate metric guidance from a PixelGoal to remain competitive with planners given metric targets directly.

The qualitative results provide further evidence. Fig.~\ref{fig:planner_qualitative_results} shows that OccPlanner generates trajectories through free space across diverse layouts. Fig.~\ref{fig:pixelgoal_comparison} follows the three PixelGoal planners on the same episode. OccPlanner updates its trajectory, passes the intervening obstacle, and continues toward the target, whereas NavDP-PixelGoal and PixNav remain blocked.

\noindent\textbf{Contributions of Goal and Occupancy Reasoning.}
We construct four cumulative variants: (1) \textbf{RGB-D Baseline}, which uses the shared RGB-D representation, PixelGoal encoder, and diffusion planner; (2) \textbf{+ Metric Goal Grounding}, which adds the ego-goal context pathway and egocentric goal supervision; (3) \textbf{+ Occupancy Supervision}, which further adds occupancy prediction without supplying its features to the planner; and (4) \textbf{OccPlanner}, which additionally supplies goal-conditioned occupancy tokens to the planner.

\begin{figure}[!t]
\centering
\includegraphics[width=\linewidth]{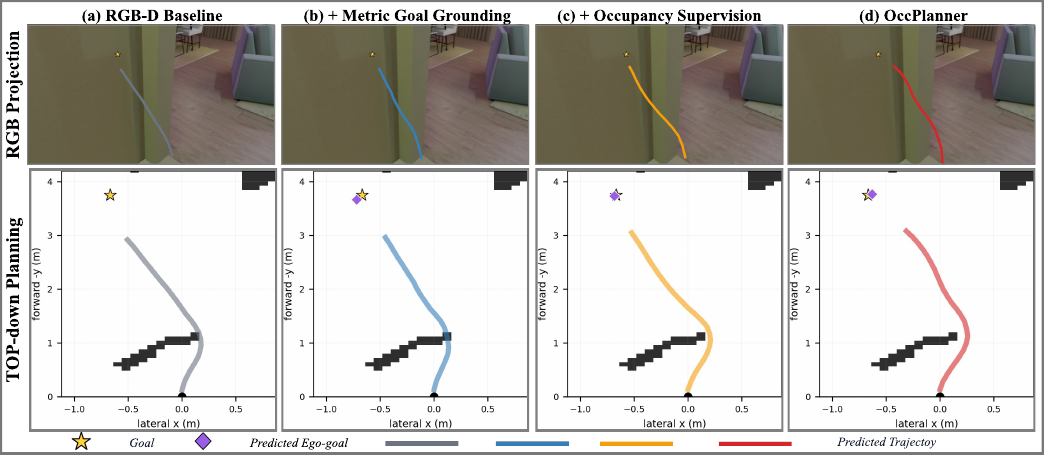}
\caption{\textbf{Qualitative ablation under target occlusion.} Goal grounding provides target direction (a--b), while occupancy supervision and conditioning progressively improve obstacle avoidance (b--d).}
\label{fig:ablation_insight}
\end{figure}

Metric goal grounding reduces ADE from 0.164\,m to 0.150\,m and improves both heading errors, although FDE increases. It guides local motion toward the target rather than constraining the final waypoint. Fig.~\ref{fig:ablation_insight} shows an occluded-target case under our reprojected PixelGoal protocol. The target lies behind a intervening obstacle. Goal grounding better aligns the trajectory with the target direction in (b), although it still grazes the occupied region. Occupancy supervision moves the trajectory outside this region in (c). Direct occupancy conditioning further increases obstacle clearance in (d) and achieves the lowest ADE and FDE of 0.130\,m and 0.135\,m. Its slightly lower visible-occupancy IoU also shows that occupancy IoU alone does not measure planning utility.

\noindent\textbf{Effect of Scene Structure and Goal Distance.}
Increasing the goal distance has a larger effect in Home and Commercial than in the cluttered scenes. From the medium to the hard split, OccPlanner's SR drops by only 0.38 points in Cluttered Easy and 2.66 points in Cluttered Hard. These scenes mainly contain obstacles placed on flat floors. Home and Commercial have more varied layouts, floor geometry, and contact properties. Rugs and low steps can cause chassis oscillation or immobilization in simulation. Longer routes increase exposure to these failures. Consequently, OccPlanner's SR drops by 14.43 and 20.93 points in Home and Commercial, and the same trend appears across the compared methods. These results capture both long-horizon planning and physical execution in heterogeneous environments. SPL remains close to SR, showing that successful trajectories remain efficient.

\begin{table}[!t]
\centering
\caption{Open-loop results of ablation experiments on 1{,}496 held-out InternData-N1 samples.}
\label{tab:ablation}
\scriptsize
\setlength{\tabcolsep}{2.4pt}
\renewcommand{\arraystretch}{1.10}
\resizebox{0.88\columnwidth}{!}{
\begin{tabular}{lccccc}
\toprule
\multirow{2}{*}{\textbf{Configuration}}
& \multicolumn{1}{c}{\textbf{Occupancy}}
& \multicolumn{4}{c}{\textbf{Trajectory Prediction}} \\
\cmidrule(lr){2-2}\cmidrule(lr){3-6}
& IoU (\%)$\uparrow$
& ADE (m)$\downarrow$
& FDE (m)$\downarrow$
& AHE ($^\circ$)$\downarrow$
& FHE ($^\circ$)$\downarrow$ \\
\midrule
RGB-D Baseline
& -- & 0.164 & 0.176 & 12.73 & 19.10 \\
+ Metric Goal Grounding
& -- & 0.150 & 0.193 & 11.06 & 16.60 \\
+ Occupancy Supervision
& \textbf{49.87} & 0.146 & 0.163 & 10.38 & 15.71 \\
\rowcolor{gray!5}
\textbf{OccPlanner}
& 47.02 & \textbf{0.130} & \textbf{0.135}
& \textbf{9.99} & \textbf{15.31} \\
\bottomrule
\end{tabular}
}
\end{table}

\begin{figure}[!t]
\centering
\includegraphics[width=0.92\linewidth]{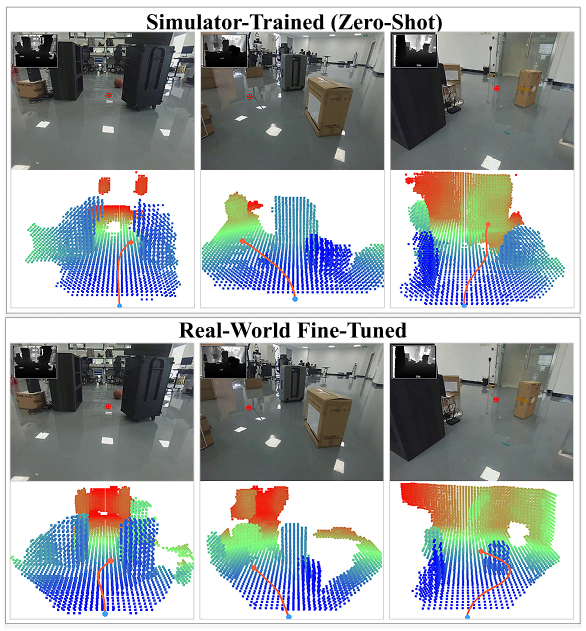}
\caption{\textbf{Real-world fine-tuning.} First-person occupancy and trajectory predictions during closed-loop deployment of the simulator-trained and real-world-fine-tuned models.}
\label{fig:real_compare}
\end{figure}

\noindent\textbf{Real-World Transfer and Closed-Loop Deployment.}
We deploy OccPlanner on a Unitree Go2 equipped with an Orbbec Gemini 336L RGB-D camera mounted 18\,cm above the robot head. Inference runs on an NVIDIA RTX 4090 GPU. SAM 3~\cite{carion2025sam3segmentconcepts} updates the PixelGoal at 1\,Hz, and OccPlanner predicts a 24-waypoint local trajectory. Dynamic-VINS~\cite{liu2022dynamicvins} estimates ego-motion at 20\,Hz. An MPC controller tracks the next six waypoints and outputs velocity commands.

When the target leaves the field of view, MPC continues tracking the remaining trajectory using odometry. If the trajectory is exhausted before redetection, the last observed target position is propagated using the accumulated ego-motion for replanning. This allows closed-loop navigation under intermittent target visibility.

We fine-tune OccPlanner on 829 real-world samples annotated by L3ROcc and evaluate both training settings over 20 trials each. A trial is successful if the robot reaches within 0.5\,m of the target. A collision is recorded when any part of the robot contacts an obstacle. Success and collision are measured independently.

\begin{figure}[!t]
\centering
\includegraphics[width=\linewidth]{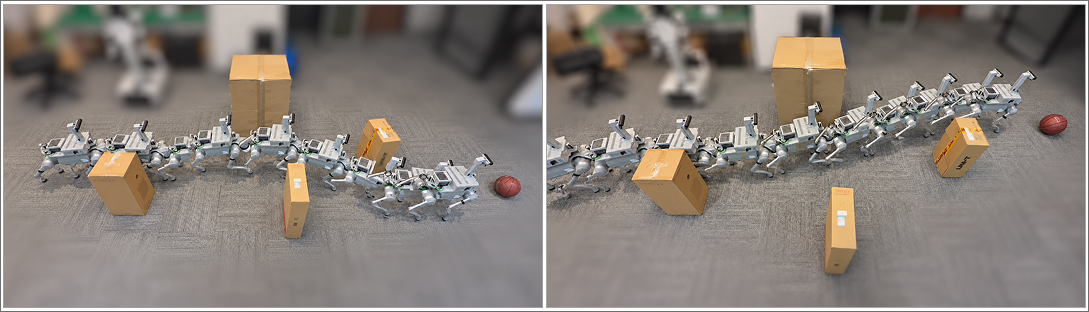}
\caption{\textbf{Real-world closed-loop navigation.} Executed robot motion during obstacle avoidance in two representative trials.}
\label{fig:real_closed_loop_trails}
\end{figure}

\begin{table}[!t]
\centering
\caption{\textbf{Real-world navigation.} Closed-loop results on Unitree Go2.}
\label{tab:real_closed_loop}
\scriptsize
\setlength{\tabcolsep}{8pt}
\renewcommand{\arraystretch}{1.10}
\begin{tabular}{lcc}
\toprule
\multirow{2}{*}{\textbf{Training Setting}}
& \multicolumn{2}{c}{\textbf{Unitree Go2 (20 Trials)}} \\
\cmidrule(lr){2-3}
& \textbf{SR}$\uparrow$ & \textbf{CR}$\downarrow$ \\
\midrule
Simulator-trained (zero-shot) & 11/20 & 12/20 \\
\rowcolor{gray!5}
\textbf{Real-world fine-tuned} & \textbf{16/20} & \textbf{5/20} \\
\bottomrule
\end{tabular}
\end{table}

As shown in Table~\ref{tab:real_closed_loop}, real-world fine-tuning increases SR from 55\% to 80\% and reduces the collision rate from 60\% to 25\%. Fig.~\ref{fig:real_compare} shows more complete and coherent obstacle predictions after fine-tuning. The resulting trajectories also follow the observed free space more closely.These results show that fine-tuning on real-world samples help adapt the learned geometry from simulation to real RGB-D observations. Fig.~\ref{fig:real_closed_loop_trails} shows the robot adjusting its path around different obstacle layouts while progressing toward the target.

Residual contacts mainly occur after the head-mounted camera passes an obstacle while the torso or legs remain beside it. The obstacle then leaves the current view. This exposes two remaining limitations: the planner has no explicit whole-body clearance constraint and no persistent occupancy memory.
\section{Conclusion}
\label{sec:conclusion}


We presented OccPlanner, a goal-aware occupancy-conditioned diffusion planner for PixelGoal navigation. OccPlanner learns an egocentric metric goal representation and extracts target-relevant geometry from occupancy to condition obstacle-aware trajectory generation. We also introduced L3ROcc, which generates aligned local occupancy and trajectory supervision from monocular RGB navigation videos.


Experiments across about 5{,}000 valid closed-loop episodes demonstrate competitive navigation performance against metric-goal planners, while ablations show the complementary effects of goal grounding and occupancy conditioning. Closed-loop deployment on a Unitree Go2 further provides preliminary evidence of real-world transfer. 

Several limitations remain. L3ROcc assumes static scenes and may incorporate moving objects into the reconstructed geometry. Motion-aware filtering could extend annotation generation to dynamic environments. OccPlanner lacks explicit whole-body clearance constraints, which may cause contacts during deployment. Its occupancy branch currently benefits substantially from depth observations, while RGB-only occupancy encoding could enable broader deployment.

\bibliographystyle{IEEEtran}
\bibliography{main}

\end{document}